\pdfoutput=1

\documentclass[11pt]{article}

\usepackage[]{EMNLP2023}

\usepackage{times}
\usepackage{latexsym}

\usepackage[T1]{fontenc}

\usepackage[utf8]{inputenc}

\usepackage{microtype}

\usepackage{inconsolata}

\usepackage{graphicx}
\graphicspath{ {./images/} }
\usepackage{multirow}
\usepackage[utf8]{inputenc} 
\usepackage[T1]{fontenc}    
\usepackage{enumitem}
\usepackage{hyperref}       
\usepackage{url}            
\usepackage{booktabs}       
\usepackage{amsfonts}       
\usepackage{nicefrac}       
\usepackage{microtype}      
\usepackage{xcolor}         
\usepackage{caption}
\usepackage{subcaption}
\usepackage{makecell}
\usepackage{amsmath}
\usepackage{mathtools}
\usepackage{amssymb}

\usepackage{titlesec}

\title{Multi-teacher Distillation for Multilingual Spelling Correction}

\author{Jingfen Zhang \\
  Amazon.com Inc\\
  \texttt{jingfenz@amazon.com} \\\And
  Xuan Guo \\
  Amazon.com Inc\\
  \texttt{xuangu@amazon.com} \\\And
  Sravan Bodapati \\
  Amazon.com Inc\\
  \texttt{sravanb@amazon.com} \\\And
  Christopher Potts \\
  Stanford University \\
  \texttt{cgpotts@stanford.edu} \\}

\begin{document}
\maketitle
\begin{abstract}
Accurate spelling correction is a critical step in modern search interfaces, especially in an era of mobile devices and speech-to-text interfaces. For services that are deployed around the world, this poses a significant challenge for multilingual NLP: spelling errors need to be caught and corrected in all languages, and even in queries that use multiple languages. In this paper, we tackle this challenge using multi-teacher distillation. On our approach, a monolingual teacher model is trained for each language/locale, and these individual models are distilled into a single multilingual student model intended to serve all languages/locales. In experiments using open-source data as well as user data from a worldwide search service, we show that this leads to highly effective spelling correction models that can meet the tight latency requirements of deployed services.
\end{abstract}

\section{Introduction}

Spelling correction is vital to the modern search experience. Users expect it, mobile devices and speech-to-text interfaces make it more crucial than ever, and uncaught spelling errors can lead to urgent problems of security and trust if problematic search results are shown to users. For services with a global reach, this poses a substantial challenge for multilingual NLP: spelling errors must be caught and corrected in any language, and even in queries using multiple languages. 

The promise of multilingual language models is that we may be able to meet these challenges with a single spelling correction model serving all languages/locales. In the present paper, we develop and motivate such a multilingual approach relying crucially on multi-teacher distillation. On our approach, an individual teacher model is trained for each language/locale, and these individual models are distilled into a single multilingual student model intended to serve all languages/locales. 

Our distillation objective is a purely behavioral one: the multilingual student is trained to mimic the input--output behavior of the individual teachers. This brings a number of key advantages in our setting. First, we can customize the individual teacher models to specific languages/locales, which proves especially useful in the area of tokenization. Second, when we want to add a new language/locale~$L$, we train just two models: the new teacher for~$L$ and the new multilingual model distilled from the input--output pairs generated by all the teacher models. Third, the individual teacher models are themselves assets that can be distilled into student models; where these are superior (common for data-rich languages), they can be used.

We motivate our approach with a wide range of experiments using open-source data as well as proprietary user data from a worldwide search service. Overall, we find that our multi-teacher distillation approach leads to superior models compared to both individual monolingual student models and multilingual student models distilled from a single multilingual teacher. In addition, we show that we can efficiently add new languages and easily meet the tight latency requirements imposed by industrial search engines. Overall, we suggest that this is a promising modeling approach not only for spelling correction but also for the other services needing to serve numerous languages and locales.

\section{Related Work}\label{sec:related_work}

Spelling correction is a widely studied problem \cite{hladek2020survey}. Earlier work relied on lexical rules \cite{meddeb1994logic, reynaert2004multilingual} or language models plus linguistic features~\cite{alkhafaji2013new, sharma2023contextual}. In more recent work, spelling correction is cast as an encoder--decoder problem \cite{hasan2015spelling, zhou2017spelling, kuznetsov2021spelling}, theoretically making it easier to scale to multilingual settings. However, methods that are efficient and scalable across numerous languages have been much less explored.



Spelling correction is highly sensitive to different tokenization schemes, since it involves manipulating characters and other subword units. Subword tokenization schemes provide the right balance between operating on subword units and being efficient for training and inference. Popular methods for subword tokenization include Byte Pair Encoding (BPE;~\cite{bostrom2020byte}), Byte-Level BPE (BBPE;~\cite{wang2020neural}), SentencePiece~\citep{kudo2018sentencepiece}, and Unigram Language Model~\citep{kudo2018subword}. BPE splits words into subword units based on their statistical properties and is extensively employed by various Transformer models such as GPT~\cite{radford2018improving}, 
RoBERTa~\cite{liu2019roberta}, and BART~\cite{lewis2019bart}. BBPE operates at the byte-level, efficiently enabling the encoding and decoding of texts across different languages with non-overlapping character sets. In this paper, we explore BPE and BBPE tokenization schemes in our experiments.

Increasing model size and amount of compute used for training will generally improve the performance of neural language models ~\cite{kaplan2020scaling}, but the costs might be prohibitive.
In model distillation \cite{hinton2015distilling}, a large teacher model is used to guide the training of a smaller student model. This is a viable solution 
for developing deployable models with strict production constraints. These approaches have proven highly successful for seq2seq problems in general \citep{kim2016sequence, liang2022multi}. Distillation approaches vary in the degree to which they presuppose access to the teacher model during student model training~\citep{gou2021knowledge}. At one extreme, the teacher and student models are trained together (i.e., co-distillation;~\citealt{chung2020feature}). At the other extreme, the teacher is simply used to generate output labels for the training data, based on which the student is trained (e.g., Sequence-level Knowledge Distillation (Seq-KD); \citealt{kim2016sequence}). In our multi-teacher distillation, we aim to decouple the teacher and student training regimes in order to train the best model for each language, and  so we use Seq-KD. There are existing studies about multiple teacher distillation. For example, You et al. combined multiple teacher networks by averaging the softened output targets and selecting layers in student and teacher networks~\citep{you2017learning}. Yuan et al. selected soft labels from a collection of teacher models, based on the reward signal from performance of distilled student model~\citep{yuan2021reinforced}. Both studies use multiple teachers to generate variant candidates and distill knowledge to build a robust and accurate student. In this paper, we apply multi-teachers for multilingual problem, where each teacher specializes in one language and they work together to guide the learning of a multi-lingual student.


\section{Methods}\label{sec:method}
\begin{figure}[tp]
\centering
\includegraphics[width=0.8\linewidth]{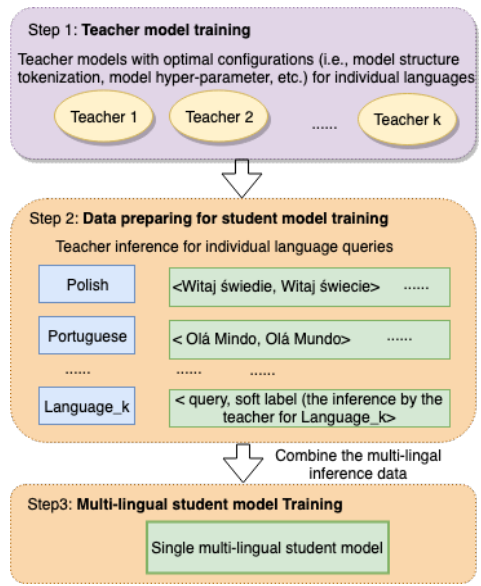}
\caption{Multi-teacher distillation workflow.}
\label{fig:multi_teacher_distillation_workflow}
\end{figure}
We aim to create a high-performing model that can serve multiple languages while satisfying latency restrictions. We propose a multi-teacher distillation method. The main idea is to train teacher models with optimal configurations for individual languages and then build a single student model based on the multi-teacher inference.  Figure~\ref{fig:multi_teacher_distillation_workflow} provides a high-level overview of our multi-teacher distillation approach, which we describe in this section.

\subsection{Model Architecture}
We first formulate the spelling correction task as a text-to-text problem: a query is the input to the  model, and the model outputs a correctly spelled query. If the model detects no spelling errors in the input, it outputs a query identical to the input. 

We use the Bidirectional Auto Regressive Transformer (BART; \citealt{bart}) architecture for model building. BART is pretrained on a denoising objective mapping corrupted sequences into their uncorrupted forms, which has many similarities to spelling correction itself. However, our approach is not specific to BART. Indeed, our very lightweight distillation objective even allows different architectures to be used for different languages.

All the spelling correction models reported in this paper are trained from scratch on spelling correction datasets rather than starting from pretrained parameters. This might seem surprising given widespread evidence that pretraining improves models. For example, multilingual BART (mBART)~\citep{liu2020multilingual} is reported as a good pretrained model for many multilingual tasks such as machine translation~\citep{maurya2021zmbart}, text generation~\citep{chen2021mtg}, text summarization~\citep{wang2021contrastive}, and entity linking~\citep{de2022multilingual}. However, spelling correction is arguably a different area from the other tasks. First, pretraining objectives tend to serve semantic goals, whereas many aspects of spelling correction are purely form-based \citep{Huang-etal:2022}. Second, spelling correction training datasets can be truly massive, since gold behavior data can be expanded with synthetic examples. As a result of these factors, the contributions of pretraining are minimal in practice. For our purposes, this has the advantage of leading to more controlled experimental comparisons, as we do not have to worry about variation in pretraining as a factor in model performance.

\subsection{Teacher Training}
For teacher training, we train different customized individual teacher models for each language to achieve high performance. For example, we adapt BPE or BBPE tokenization methods according to each language's characteristics, and build both monolingual and multilingual models with different hyper-parameters based on  language difficulty and training data availability. The optimal choice varies between languages, and our approach can accommodate this in the teacher creation phase.

\subsection{Distillation Objective}\label{sec:distill}



As discussed in Section~\ref{sec:related_work}, we use the Seq-KD method of \citet{kim2016sequence}, in which the teacher is simply used to generate ``soft'' labels for student training. This has led to exceptional spelling correction models in practice, in the context we describe in Section~\ref{sec:experiments:user}. In addition, it is extremely efficient in terms of overall system development, and it allows the teacher and student models to have different sizes, tokenization schemes, and other architectural features.

In our experiments, we explore a range of options: (1) a multilingual teacher distilled into a multilingual student model; (2) multi-teacher distillation using each of the monolingual teachers; and (3) multi-teacher distillation from the best teacher for each language, selected from the set of monolingual and multilingual teachers available. It turns out that the option (3) provides the best results.

A guiding hypothesis for our method is that our distillation process can lead to individual models that are not only capable of serving all languages/locales, but also superior to monolingual models due to knowledge sharing across languages. We expect to see the largest gains in low-resource languages, and this is indeed what we find experimentally.

\subsection{Evaluation Metric}

Our evaluations are based in exact match (after punctuation removal) between gold and predicted outputs, and we focus on cases involving corrections to avoid inflating our scores with inputs that contain no spelling errors. Thus, precision is the percentage of model-predicted corrections that are correct according to the gold data, and recall is the percentage of cases requiring corrections that the model predicts correctly. We report the F1 score of these two values. Appendix~\ref{sec:evaluation_metrics} provides additional details on score calculation.


\section{Experiments on User Data}\label{sec:experiments:user}
In this section, we report on experiments with user data from a large, global search service, and the user data are their search queries. In Section~\ref{sec:open-source-data}, we report on experiments with open source data that are natural language sentences. With the open-source data, we can be completely open about the findings, with some costs in terms of realism. With the user data, we are required to conceal some details, but the findings themselves still provide a clear picture of how our approach fares in the real world.

\begin{table*}[tp]
\begin{minipage}{\textwidth}
\centering
\resizebox{1\linewidth}{!}{
\begin{tabular}{l l c rrrr c rr c rr}
\toprule
&& 
& \multicolumn{4}{c}{Teachers} 
&& \multicolumn{2}{c}{Single-teacher distillation}
&& \multicolumn{2}{c}{Multi-teacher distillation}
\\
&&
& \multicolumn{1}{c}{Multi-BPE \footnote{The baseline model.}}
& \multicolumn{1}{c}{Multi-BBPE}
& \multicolumn{1}{c}{Mono-BBPE}
& \multicolumn{1}{c}{Best}
&& \multicolumn{1}{c}{Multi-BPE}
& \multicolumn{1}{c}{Multi-BBPE}
&& \multicolumn{1}{c}{Matched mono}
& \multicolumn{1}{c}{Best} 
\\
Language
& Locale
&& \multicolumn{1}{c}{teacher}
& \multicolumn{1}{c}{teacher}
& \multicolumn{1}{c}{teacher}
& \multicolumn{1}{c}{teacher}
&& \multicolumn{1}{c}{student}
& \multicolumn{1}{c}{student}
&& \multicolumn{1}{c}{student} 
& \multicolumn{1}{c}{student} \\
\cmidrule{1-2}  \cmidrule{4-7} \cmidrule{9-10} \cmidrule{12-13}
Portuguese & BR && $-$ & $-$1.2\% &  3.3\%	& 3.3\%	 && $-$3.7\%	& $-$2.8\%	&& $-$2.9\%	& $-$1.1\% \\
Dutch      & NL && $-$ &  5.5\% & $-$5.2\%	& 5.5\% && $-$0.6\%	& $-$0.3\%	&&  0.9\%	&  3.4\% \\
Turkish    & TR && {\bf $-$} & $-$1.1\% & $-$2.0\%	& 0	     && $-$0.4\%	& $-$6.6\%	&& $-$3.2\%	&  0.8\%\\
Swedish    & SE && $-$ &  1.1\% & $-$6.0\%	& 1.1\%	 && $-$4.7\%	& $-$1.2\%	&&  3.3\% &  2.0\%\\
Polish     & PL && {\bf $-$} &	$-$4.0\% & $-$0.5\%	& 0	     && $-$5.3\%	&  1.6\%	&&  2.5\%    &  7.2\%\\
           & AE && $-$ &  6.3\% & 12.5\% & 12.5\% && $-$0.4\%	& $-$0.6\%	&& 14.3\%	& 20.4\%\\
Arabic     & SA && $-$ &  8.7\% & 16.6\%	& 16.6\% && $-$0.1\%	&  6.5\%	&& 25.1\%	& 28.5\%\\
           & EG && $-$ &  4.9\% & 16.5\%	& 16.5\% &&	1.0\%	&  1.3\%	&& 23.1\%	& 32.5\%\\
\cmidrule{1-2}  \cmidrule{4-7} \cmidrule{9-10} \cmidrule{12-13}
\multicolumn{2}{l}{Avg across locales} && $-$ & 1.7\% &	2.2\%	& 5.2\%  &&	$-$2.2\%	& $-$1.0\% &&	4.7\%	& 8.0\%\\
\hline
\end{tabular}}
\caption{F1 scores on user data. Due to external constraints, we report only percentage-wise changes relative to the Multi-BPE model, whose absolute performance we cannot disclose. The multi-teacher students (far right two columns) yield the best results. Here, `Matched mono' is the multilingual model distilled from the column of models represented under `Monolingual teachers', whereas `Best' is the the multilingual model distilled from the column of models represented under `Best teachers'. Overall, these results indicate that multi-teacher distillation is an effective strategy for industrial spelling correction, and that the flexibility afforded by our lightweight distillation strategy pays off.}
\label{tab:model_performance}
\end{minipage}
\end{table*}

\subsection{Training Data} 

Our user data are derived from a global search service. For a proof of concept, we focus on six languages: Portuguese, Dutch, Turkish, Swedish, Polish, and Arabic. These cover eight locales: Brazil~(BR), Netherlands~(NL), Turkey~(TR), Sweden~(SE), Poland~(PL), United Arab Emirates~(AE), Saudi Arabia~(SA), and Egypt~(EG). We collected two years of historical behavior data (2021 to 2023), comprising <input query, label query> pairs. In this context, the input query refers to the user's initial search query, while the label query represents the prediction made by our production speller model as validated by user data (successful completion of a search as measured by clicks and other behavior). We have millions to hundreds of millions of examples, with imbalanced size across locales. For example, the data for PL is less than 1/20 the size of BR. 
 
\subsection{Evaluation Data} 

We collected human annotations of search queries to serve as ground truth in model evaluations. For each locale, there are 10K human-annotated queries that reflect the spelling correction distribution in production. These examples are collected from a different time window than the training data collection, and they are carefully sampled to balance cases where misspellings could have been corrected and where good spellings should not have been over-corrected.

\subsection{Monolingual Teachers} 

The first step of our method is to train individual teacher models, including both monolingual and multilingual models with optimal configurations to reach high performance in each language. A major advantage of our approach is that we can train a diverse set of models, using choices that are tightly aligned with what we know about individual languages. We heuristically explored different configurations for different languages. This led us to use a full-size BART-large model with a 128K BPE vocabulary (480M parameters) for BR, and a 6-layer BART model with 32K BBPE vocabulary (211M parameters) for the other locales.

\subsection{Multilingual Teachers}  

We trained two multilingual teacher models with the full-size BART-large architecture. The \textbf{Multi-BPE} model uses BPE tokenization and has a 128K vocabulary (490M parameters). This model serves as a baseline for all our comparative reporting. The \textbf{Multi-BBPE} model uses BBPE tokenization and has a 32K vocabulary (471M parameters).

\subsection{Multi-Teacher Distillation}

We distill teacher models into student models according to the methods described in Section~\ref{sec:distill}. All student models are BART-base models with 2~layers, trained with 25 training epochs. Each epoch contains 200 millions of training data. These models are small compared to the teacher model due to our latency requirements (Section~\ref{sec:latency}).

\subsection{Results} 

Our results are summarized in Table~\ref{tab:model_performance}. Due to external constraints, we can show only percentage-wise gains and losses relative to the Multi-BPE teacher model, rather than reporting raw F1 scores. Nonetheless, the findings are very clear: our multi-teacher distillation approach is superior, leading to solid gains in nearly every locale and a very large average improvement across locales. The best students are those distilled from the best teacher for each language (rightmost column).

Some variation is observed across different languages and locales. For example, a significant difference of 8.7\% is observed in SA and a 5.5\% difference is observed in NL. When comparing the monolingual models with the multilingual models, a similar pattern is observed, with comparable overall performance but even larger variation across languages and locales. On our approach, we can embrace this variation and choose the best teacher for each language to obtain a better multi-lingual student model. 

Multi-teacher distillation out-performs the corresponding monolingual teacher in all locales except BR. BR is the largest of these locales, and it is common for large locales to support very strong monolingual models; the strengths of multi-teacher distillation are usually most apparent in low-resource locales. During  multilingual student training, we observed differences across languages. While the training for most languages achieved convergence, the training on Portuguese data did not converge optimally. Although we could achieve better performance on Portuguese by doing more training, over-fitting could result in a sacrifice of performance on the other languages. In the future, we plan to address this by treating different languages as different tasks and developing a multi-task approach that dynamically allocates computing effort to different languages~\citep{ruder2017overview,duong2015low,baxter1997bayesian}.

\subsection{Adding New Languages}

The results in Table~\ref{tab:model_performance} shows that multilingual student training has the capacity to transfer knowledge among languages. In addition, the approach makes it easy to include new languages or data in the future with minimal effort: we simply add the new monolingual teacher model inference data into the distillation process and expand the multilingual student model without having to retrain the entire set of teacher models for all languages from scratch. 

To illustrate, we trained a multilingual student model using monolingual teacher inference data from three languages: Portuguese, Dutch, and Polish. We obtained an improvement of 4.7\% in the average F1 score of the student model compared to the average F1 score of the teachers. We then added two more languages (Turkish and Swedish) and obtained similar F1 scores for Portuguese, Dutch, and Polish while achieving better performance for Turkish and Swedish than their respective monolingual teachers. Table~\ref{tab:add_new_lang} summarizes these experiments.

\begin{table}[!t]
\centering
\resizebox{1\linewidth}{!}{
\begin{tabular}{l c r r}
\toprule
Locale 
& \multicolumn{1}{c}{Monolingual}
& \multicolumn{1}{c}{Distill on}
& \multicolumn{1}{c}{Distill on} 
\\
& \multicolumn{1}{c}{teacher}
& \multicolumn{1}{c}{3 languages}
& \multicolumn{1}{c}{5 languages} 
\\
\midrule
BR	& $-$ &  $-$2.0\% &	$-$2.2\%\\
NL	& $-$ &	9.0\% &	9.2\% \\
PL	& $-$ &  10.0\%&	9.0\% \\
TR	& $-$ & &		4.5\%\\
SE	& $-$ & &		20.6\% \\
\midrule
Avg (3)	&	$-$ & 4.7\% &	4.4\%\\
Avg (5)	&	$-$ & &		7.0\%\\
\bottomrule
\end{tabular}}
\caption{Model performance (F1) after adding new languages to the multi-teacher distillation process.}
\label{tab:add_new_lang}
\end{table}


\subsection{Latency}\label{sec:latency}

Industrial search technologies operate under very tight latency requirements. We have demonstrated that our multi-teacher distilled student model outperforms the larger teacher models (Table~\ref{tab:model_performance}), but we have not so far quantified the latency gains that this brings. 

In this section, we evaluate the impact of multi-teacher distillation on online deployment by conducting a real traffic load test to measure throughput per second (TPS) and P99 latency. A higher TPS enables a reduction in the number of GPU instances needed to handle the same volume of service requests, thereby lowering the overall Initial Margin Requirement (IMR) costs of the inference fleet. In addition, improvements in the P99 latency will allow for more spelling corrections that would otherwise result in ``no corrections'' due to timeouts. This ensures that the online F1 performance is consistent with the offline F1, leading to a better user experience.

Table~\ref{tab:Latency_numbers} lists the comparison of TP99 latency and TPS by the multi-teacher distilled student model on six locales against the multilingual teacher model reported in Table 1. For these experiments, we first convert the model to an ONNX (Open Neural Network eXchange) model graph~(\citealt{onnx2023ai}) and then optimize the serialized ONNX graph using TensorRTframework~\citep{vanholder2016efficient}. Here, all latency numbers are based on the TensorRT serialized model on an AWS g5.2xlarge GPU instance. We observe that the TPS of the student model is double that of the teacher model, and so we can save more than half of IMR costs by deploying the student models.

\begin{table}[!t]
\centering
\begin{tabular}{l c c l c c}
\toprule
 & P99 & TPS & &  P99 & TPS\\
\midrule
 BR  & 37.9\% & +2.1x  & TR  & 37.1\% & +2.1x\\
 PL  & 43.7\% & +2.1x  & SE  & 33.9\% & +2.3x \\
 NL  & 31.6\% & +2.3x  & AE  & 50.1\% & +2.6x \\
\bottomrule
\end{tabular}
\caption{Improvement in TP99 latency and throughput for the student model vs.~the baseline teacher model shown in Table~\ref{tab:model_performance}.}
\label{tab:Latency_numbers}
\end{table}

\section{Experiments on Open-Source Data}\label{sec:open-source-data}

To supplement our experiments on user data, we also conducted experiments with open-source data for which we can supply absolute performance numbers. 

\subsection{Data} 

A few spelling datasets have been proposed  \cite{hagiwara2019github, rothe2021simple}, but most of these focus primarily on English. In this paper we use the large multilingual dataset from the Workshop on Statistical Machine Translation (WMT) website.\footnote{\url{https://www.statmt.org/wmt19/translation-task.html}} We downloaded the europarl, news-commentary, and news- 2007 to news-2011 corpora for five languages, English (EN), Germany (DE), Czech(CS), French(FR) and Spanish(ES).  We then injected synthetic noise into these sequence to get <noise inserted sequence, original sequence> pairs as our training data. The operations used in noise injection include inserting, deleting, and replacing random characters at random locations. For each original sentence, we generated 8 noised sentences for training set, and one noised sentence for evaluation set. For evaluation data, we filtered out trivial cases and sequence less than 6 words, and then randomly selected 10,000 as the evaluation data for each language. Table~\ref{tab:Training and evaluation dataset} provides an overview of these resources.

\begin{table}[ht]
\centering
\begin{tabular}{c r c c}
\toprule
Language & \multicolumn{1}{c}{Train} & Eval & Overlap \\
\midrule
EN &	181,597,816	& 10,000	& 393 \\
DE &	133,116,472	& 10,000	& 418  \\
CS &	71,469,552	& 10,000	& 475 \\
FR &	47,164,952	& 10,000	& 480 \\
ES &	9,215,136	& 10,000	& 515 \\
\bottomrule
\end{tabular}
\caption{Open source data: training and evaluation data size. The overlap between the evaluation data and the training data ranges from 3.93\% to 5.15\% (denominator is evaluation size).}
\label{tab:Training and evaluation dataset}
\end{table}

\begin{table}[tp]
\setlength{\tabcolsep}{5pt}
\centering
\begin{tabular}{l c  cc  c ccc }
\toprule

&& \multicolumn{2}{c}{Teachers} 
&& \multicolumn{3}{c}{Students}\\
&& {Multi}
& {Mono}
&& S-T
& M-T
& B-T
\\
\cmidrule{1-1}  \cmidrule{3-4}  \cmidrule{6-8}
EN &&	76.0 &	77.4 &&	 71.6 &	72.9 &	73.0 \\
DE &&	90.3 &	92.0 &&	 85.2 &	87.4 &	87.5 \\
CS &&	85.0 &	85.3 &&	 77.7 &	80.2 &	80.7 \\
FR &&	44.8 &	44.4 &&	 42.2 &	42.8 &	43.0 \\
ES &&	85.1 &	81.9 &&	 81.4 &	81.0 &	82.6 \\
\cmidrule{1-2}  \cmidrule{3-4}  \cmidrule{6-8}
Avg. &&	76.3 &	76.2 && 71.6 &	72.9 &	73.4 \\
\bottomrule
\end{tabular}
\caption{Open-source data experiment results (F1 scores). Here, `Multi' and `Mono' are multilingual and monolingual teacher models, whereas `S-T', `M-T' and `B-T' are  distilled from the single multilingual teacher, multi-monolingual teachers and the best teachers, respectively. Our multi-teacher distillation approach is superior for all languages, with the best results emerging where the best teacher for each language is used.}
\label{tab:Experiments-on-open-source-data}

\end{table}

\subsection{Models} 

We conduct both monolingual and multilingual teacher training. For all teacher models, we use a BART-large model structure with 6-layer Transformers and a 32K BBPE vocabulary. As before, we compare three student models: (1) a model distilled from the single multilingual teacher, (2) a model distilled from the monolingual teachers, and (3) a model distilled from the best teacher for each language, which can be either the monolingual model for that language or the multilingual teacher.

\subsection{Results} 

Table~\ref{tab:Experiments-on-open-source-data} summarizes our findings. In terms of performance, the student model distilled from the multi-monolingual teachers outperforms the student model distilled from the single multilingual model, achieving an F1 score of 72.9 versus 71.6. The student model distilled from the best teachers surpasses both, achieving the highest F1 score of 73.4. Detailed F1 scores for different models are listed in Table ~\ref{tab:Experiments-on-open-source-data}. Note that the training data size and training epochs for different methods are equivalent, to make sure that the performance differences do not trace to these factors.

\section{Conclusion}

We developed and motivated a multi-teacher distillation approach for multilingual spelling correction. On our approach, teacher models for individual languages are used to distill a single multilingual student model. By focusing on improving the performance of teacher models for specific languages, we can enhance the overall performance of the student model. Additionally, our approach allows for the inclusion of new monolingual teacher model inference data into the distillation process, enabling the expansion of the multilingual student model without the need to retrain the entire set of teacher models for all languages. We believe that this modeling approach holds promise not only for spelling correction services but also for other services needing to serve numerous languages and locales.

\section*{Ethics Statement}
We hereby acknowledge that all of the co-authors of this work are aware of the provided ACL Code of Ethics and honor the code of conduct.

In this paper, we are focused on situations involving people from diverse linguistic and cultural backgrounds, spread all around the world. This is a very challenging context for any NLP system, and it raises the concern that models might be overfit to specific groups (usually the largest and most influential) at the expense of other groups. We certainly do not claim to have solved this problem, but we do view our proposed approach as an attempt to make cautious progress here. In particular, since we train a monolingual model for every language/locale, we can always fall back to that model if the multilingual one shows problematic transfer that degrades performance. On the other hand, we expect that, on average, the multilingual models will help to make up for data scarcity problems for specific languages, which improves the experiences of those users on our site, and that they will also allow users the freedom to use multilingual queries if they wish. Also, considering the popularity and relevance of our service, we anticipate that over time, our traffic will attract individuals from diverse linguistic and cultural backgrounds, thereby partially mitigating the issue.



%

\bibliography{anthology}
\bibliographystyle{acl_natbib}

\newpage
\clearpage
\appendix

\section*{Supplementary Materials}
\section{Evaluation Metrics}\label{sec:evaluation_metrics}
We use precision and recall as the offline spelling correction performance metrics, defined as follows:
\begin{itemize}
\item \emph{Exact match}(*): String identity after punctuation removal (e.g.,  ``women's'' and ``womens'' as equal).
\end{itemize}

\begin{equation*}
{\footnotesize
\begin{aligned}
\text{precision}&=\frac{\text{action}_e=\text{action}_s=\text{AUTO}\wedge\text{query}_e\simeq\text{query}_s}{\text{action}_s=\text{AUTO}} \\
\text{recall}&=\frac{\text{action}_e=\text{action}_s=\text{AUTO}\wedge\text{query}_e\simeq\text{query}_s}{\text{action}_e=\text{AUTO}}
\end{aligned}
}
\end{equation*}

\begin{itemize}[leftmargin=*,noitemsep,topsep=0pt]
\item Subscript $s$: the model output.
\item Subscript $e$: the gold (human-judged) output.
\item action: the suggested action. The possible values are \textsc{auto} (auto correction) and \textsc{none} (no correction). 
\item query: the corrected query in the case of auto correction  
\item query$\_s$ $\simeq$ query$\_e$: query$\_s$ is an exact match with query$\_e$.

\end{itemize}

\end{document}